%% file: main.tex
\documentclass{article}
\usepackage{ijcai25}
\usepackage{times}
\usepackage{soul}
\usepackage{url}
\usepackage[utf8]{inputenc}
\usepackage[small]{caption}
\usepackage{graphicx}
\usepackage{amsmath}
\usepackage{amsthm}
\usepackage{booktabs}
\usepackage{algorithm}
\usepackage{algorithmic}
\usepackage[switch]{lineno}

\usepackage{epsfig}
\usepackage{amssymb}
\usepackage{multicol}
\usepackage{multirow}
\usepackage{rotating}
\usepackage{booktabs}
\usepackage{wrapfig}
\usepackage{colortbl}

\definecolor{cvprblue}{rgb}{0.21,0.49,0.74}
\definecolor{mygray}{gray}{.9}
\definecolor{myblue}{RGB}{0,173,238}
\definecolor{myred}{RGB}{255,0,0}
\usepackage{bm}
\usepackage{wasysym}
\usepackage{utfsym}
\usepackage{diagbox}

\usepackage{xspace}

\makeatletter
\newcommand{\thickhline}{%
\noalign {\ifnum 0=`}\fi \hrule height 1pt
\futurelet \reserved@a \@xhline
}

\usepackage[pagebackref,breaklinks,colorlinks,citecolor=cvprblue]{hyperref}

\makeatletter
\DeclareRobustCommand\onedot{\futurelet\@let@token\@onedot}
\def\@onedot{\ifx\@let@token.\else.\null\fi\xspace}

\def\eg{\emph{e.g}\onedot} 
\def\ie{\emph{i.e}\onedot}

\makeatother

\title{Prompt-Aware Controllable Shadow Removal}

\author{
Kerui Chen
\and
Zhiliang Wu
\and
Wenjin Hou
\and
Kun Li
\and
Hehe Fan
\and
Yi Yang\\
\affiliations
ReLER Lab, CCAI, Zhejiang University\\
}

\begin{document}

\maketitle

\begin{abstract}
Shadow removal aims to restore the image content in shadowed regions. 
While deep learning-based methods have shown promising results, they still face key challenges: 1) uncontrolled removal of all shadows, or 2) controllable removal but heavily relies on precise shadow region masks.
To address these issues, we introduce a novel paradigm: prompt-aware controllable shadow removal. Unlike existing approaches, our paradigm allows for targeted shadow removal from specific subjects based on user prompts (e.g., dots, lines, or subject masks). 
This approach eliminates the need for shadow annotations and offers flexible, user-controlled shadow removal. 
Specifically, we propose an end-to-end learnable model, the \emph{\textbf{P}}rompt-\emph{\textbf{A}}ware \emph{\textbf{C}}ntrollable \emph{\textbf{S}}hadow \emph{\textbf{R}}emoval \emph{\textbf{Net}}work (PACSRNet).
PACSRNet consists of two key modules: a prompt-aware module that generates shadow masks for the specified subject based on the user prompt, and a shadow removal module that uses the shadow prior from the first module to restore the content in the shadowed regions. 
Additionally, we enhance the shadow removal module by incorporating feature information from the prompt-aware module through a linear operation, providing prompt-guided support for shadow removal. 
Recognizing that existing shadow removal datasets lack diverse user prompts, we contribute a new dataset specifically designed for prompt-based controllable shadow removal. 
Extensive experimental results demonstrate the effectiveness and superiority of PACSRNet.
\end{abstract}

\section{Introduction}
Shadow removal is a fundamental visual restoration task, which aims to restore the information of the darkness region caused by light occlusion in an image~\cite{liu2024recasting,vasluianu2024ntire}.
Realistic shadow removal can benefit various computer vision tasks, such as object segmentation~\cite{yang2022decoupling,luo2024soc}, object tracking~\cite{meinhardt2022trackformer,wu2024single}, face recognition~\cite{du2022elements,kim2022adaface}, content completion~\cite{10222097,wu2024waveformer}, and so on.
Therefore, shadow removal methods have been extensively studied in computer vision.
With rapid development of deep learning and increasing interest in image editing~\cite{alaluf2022hyperstyle} and augmented virtual reality~\cite{zhang2024advances}, shadow removal are getting increased attention in recent years.

\begin{figure}[!t]
\centering
\includegraphics[width=1.0\linewidth]{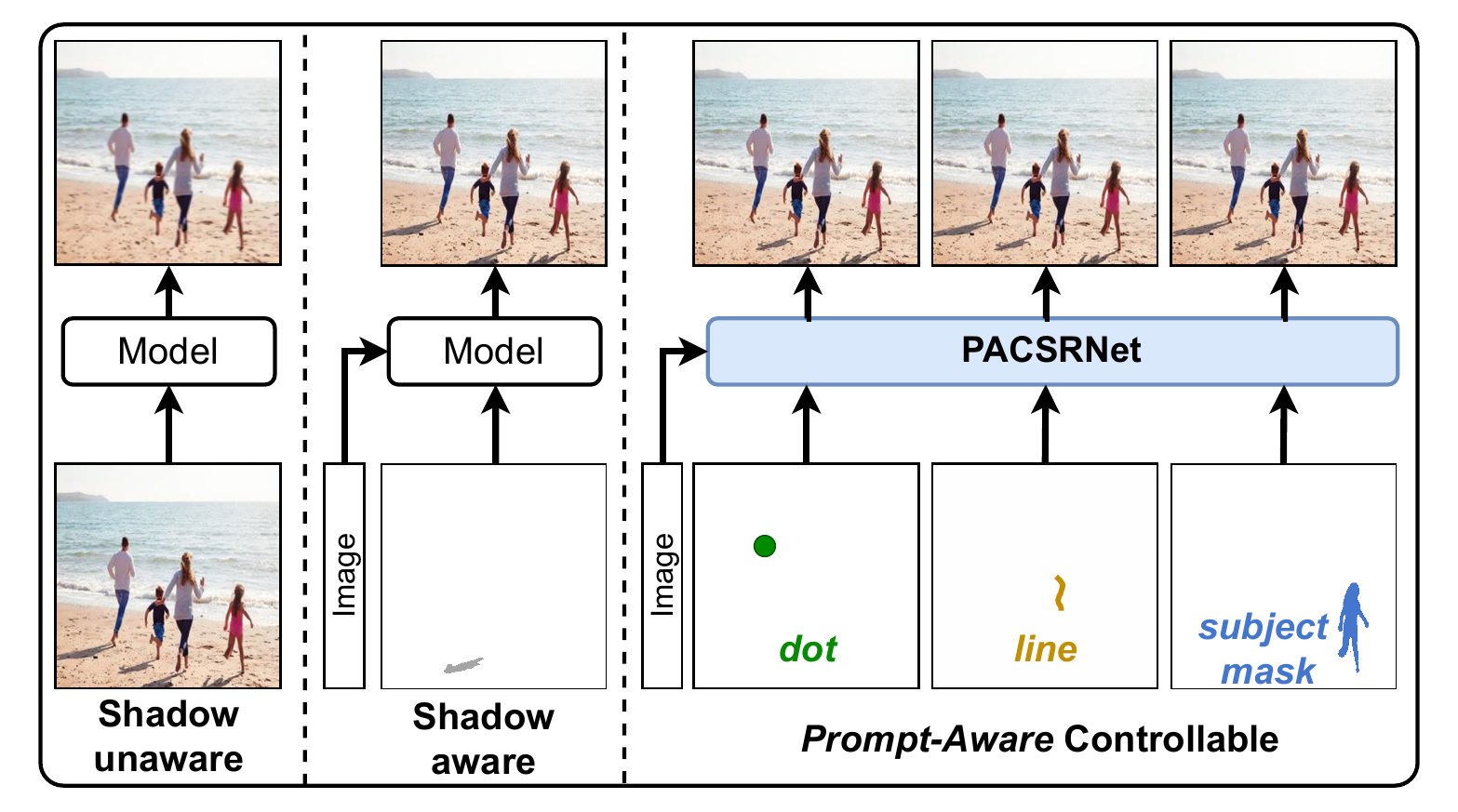}
\caption{Comparison with existing shadow removal methods. (a) \textbf{Shadow unaware} removes all the shadow regions with only raw images as input. (b) \textbf{Shadow aware} removes the shadow regions corresponding to the given shadow mask. (c) Our \textbf{prompt-aware removal}. Different from (a) and (b), the proposed method allows the removal of any subject's shadow with various prompts (\ie, dot, line, and subject mask). 
}
\label{fig:intro}
\end{figure}

Recently, several deep learning-based shadow removal methods have been proposed. 
These methods generally follow two paradigms: global shadow removal without shadow mask~\cite{hu2019direction,wang2024softshadow,li2024controlling}, and shadow removal requiring a precise shadow mask as prior information~\cite{guo2023shadowformer,guo2023shadowdiffusion,xiao2024homoformer}. 
The former is referred to as \emph{shadow unaware}, while the latter is called \emph{shadow aware}. 
Although these methods have shown promising results, they still suffer significant challenges in practical applications.
On the one hand, shadow unaware methods aim to globally remove shadows from images in a coarse-grained manner, as shown in Fig.~\ref{fig:intro} (a). 
They lack the ability to perform fine-grained removal based on user-specific needs, making them uncontrolled.
On the other hand, shadow aware methods allow for control over the removed shadows, but they rely on manually annotated masks of shadow region, as shown in Fig.~\ref{fig:intro} (b). 
However, manually annotating precise shadow regions is non-trivial due to the blur boundary between shadow and non-shadow regions, and it is time-consuming and labor-intensive.

In this paper, we propose a new paradigm, named \emph{\textbf{prompt-aware controllable shadow removal}}, to address the aforementioned challenges. 
Unlike existing shadow unaware methods~\cite{hu2019direction,wang2024softshadow,li2024controlling} and shadow aware methods~\cite{guo2023shadowformer,guo2023shadowdiffusion,xiao2024homoformer}, our proposed paradigm enables the removal of shadows from specified subjects based on user prompts, such as dot, line, and subject mask in Fig.~\ref{fig:intro} (c).
In this way, we not only achieve controllable shadow removal, but also avoid the non-trivial and tedious tasks of shadow region annotation.
These advantages ensure the flexibility and convenience of our paradigm in practical applications.

Specifically, we propose an end-to-end prompt-aware controllable shadow removal network called PACSRNet. 
It consists of a prompt-aware module and a shadow removal module.
The former aims to generate the shadow masks for a specified subject based on the user prompt, while the latter leverages the shadow prior obtained from the former to restore the contents in the shadow regions.
We further leverage the feature from the prompt-aware module via a linear operation to provide prompt-aware guidance for the shadow removal module. 
Among prompt-aware module and a shadow removal module, we design spatial-frequency interaction and dense-sparse local attention blocks as their basic units, respectively. 
The spatial-frequency interaction block aims to facilitate information interaction between spatial features and frequency features, thereby effectively enhancing the shadow perception and contextual understanding capabilities of the prompt-aware module.
The dense-sparse local attention block is used to suppress negative influence of irrelevant pixels, maximizing the utilization of relevant pixels to improve shadow removal performance.

Furthermore, existing shadow removal datasets~\cite{qu2017deshadownet,wang2018STCGAN,le2019shadow} typically contain only shadow images, shadow-free images, and shadow masks, with a lack of diverse user prompts. 
As a result, they are not suitable for the proposed prompt-aware controllable shadow removal task. 
In this paper, we introduce the first prompt-based controllable shadow removal dataset, which provides multiple user prompts for each sample, such as dot, line, and subject mask, effectively simulating real-world shadow removal scenarios.
Extensive experimental results demonstrate the effectiveness and superiority of the proposed PACSRNet.

To sum up, our contributions are summarized as follows:

\begin{itemize}
\item 
We propose a new paradigm for shadow removal: prompt-based controllable shadow removal. 
It enables the removal of shadows from specified subjects based on user prompts,
effectively alleviating the need for shadow masks.
To the best of our knowledge, this is the first work to explore prompt-based shadow removal.
\item We develop a prompt-aware controllable shadow removal network.
This network consists of a prompt-aware module and a shadow removal module, with spatial-frequency interaction and dense-sparse local attention blocks as their basic units.
\item We customize the first prompt-based controllable shadow removal dataset.     
This dataset consists of 11,900 shadow and shadow-free samples, along with various prompts (\eg, dot, line, and subject mask).
We will release it to facilitate subsequent research.
\end{itemize}

\section{Related work}
\subsection{Shadow Removal}
Classic shadow removal methods \cite{finlayson2005removal,shor2008shadow,finlayson2009entropy,yang2012shadow} typically utilized hand-crafted prior knowledge, such as illumination, region-based characteristics, and density, to remove shadow regions. 
However, such prior knowledge often lacks high-level semantic features, 
which results in the above methods usually failing in complex scenarios.
Recently, learning-based approaches~\cite{wang2018STCGAN,zhu2022bijective,li2023leveraging,guo2023shadowdiffusion,guo2023shadowformer,wang2024softshadow,mei2024latent} have significantly advanced removal effects using large-scale datasets. 
On the one hand, \cite{wang2018STCGAN,wang2024softshadow} realize end-to-end shadow unaware removal of all shadow regions. 
However, these approaches are uncontrolled and cannot target specified shadow regions for shadow removal. 
On the other hand, some studies \cite{li2023leveraging,guo2023shadowformer,guo2023shadowdiffusion} present a shadow aware removal network removing shadows by global shadow masks as prior input. 
Meanwhile, ~\cite{mei2024latent} achieves instance-level removal by using a single object's shadow mask. 
Despite achieving impressive results, these shadow aware networks still heavily rely on accurate shadow masks to indicate removal regions, which can be cumbersome and unfriendly in practical applications. 
To achieve controllable shadow removal and enhance the user experience, we propose a new prompt-aware shadow removal network that can accept diverse prompts (\ie, dot, line and subject mask), offering a more flexible and intuitive interface.

\begin{figure*}[t!]
\centering
\includegraphics[scale=0.88]{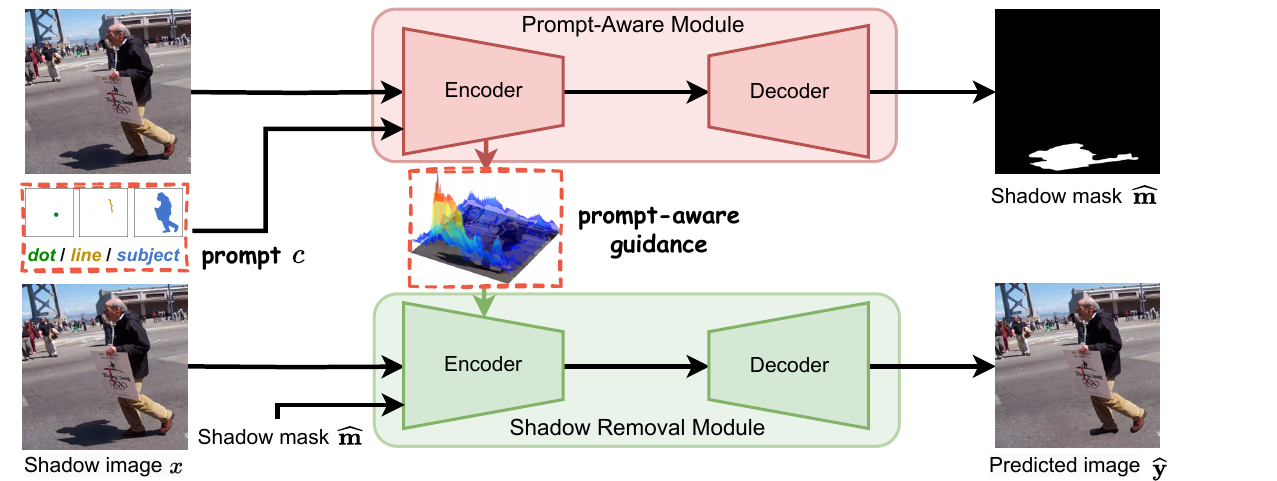}
\vspace{-0.2cm}
\caption{\textbf{The overview of the our PACSRNet.} PACSRNet is composed of a prompt-aware module and a shadow removal module. Prompt-aware module takes a shadow image and a prompt $\textbf{\emph{c}}$ as inputs generating corresponding shadow mask  $\widehat{\textbf{\emph{m}}}$ and prompt-aware guidance feature. The generated shadow mask will be served as explicit guidance and fed into the shadow removal module along with the shadow image for shadow removal. The prompt-aware guidance feature is applied to encoder of shadow removal module guiding the shadow removal implicitly.}
\label{fig2}
\vspace{-0.2cm}
\end{figure*}

\subsection{Shadow Detection}
Shadow detection~\cite{huang2011characterizes,vicente2017leave,khan2014automatic,chen2020multi} aims to identify and segment shadow regions within a given image. 
Early studies \cite{huang2011characterizes,vicente2015leave,shen2015shadow,vicente2017leave} focus mainly on utilizing various hand-crafted heuristic cues, such as color priors, image texture, or chromaticity, to identify shadow regions.  
However, when dealing with complex shadow scenes, hand-crafted features often struggle to provide adequate information and are limited in their ability to describe shadows, leading to significant degradation. 
Recently, deep learning-based methods~\cite{khan2014automatic,le2018a+,chen2020multi,liao2021shadow,yang2023silt} have achieved impressive results and have become mainstream.
For detecting all the shadow regions in the given image more accurately, \cite{khan2014automatic} first uses CNNs to learn semantic features for shadow detection, \cite{chen2020multi} presents a semi-supervised shadow detection algorithm by exploring unlabeled data through a multitask mean teacher framework, \cite{liao2021shadow} introduces the confidence map prediction network to combine the prediction results by multiple methods, and the latest method \cite{yang2023silt} uses iterative label tuning to refine noisy labels helping the network better recognize non-shadow regions and alleviate overfitting.
Although these methods can efficiently detect global shadow regions, they lack attention to controllability.
Therefore, \cite{wang2020instance,wang2021single,wang2022instance} shift to instance-level shadow detection, focusing on a finer granularity by identifying shadow regions corresponding to a specific object. 
However, these methods require an accurate object mask for shadow detection. 
In this paper, we attempt to simultaneously perform shadow removal and detection within diverse and flexible prompts.

\section{Method}

\subsection{Problem Formulation and Overview}

\noindent\textbf{{Problem Formulation.}}
Given a shadow image $\textbf{\emph{x}} \in \mathbb{R}^{h\times w\times 3}$ with width $w$ and height $h$. 
The goal of shadow removal is to generate shadow-free image $\widehat{\textbf{\emph{y}}} \in \mathbb{R}^{h\times w\times 3}$, which should be spatially consistent with the ground truth $\textbf{\emph{y}} \in \mathbb{R}^{h\times w\times 3}$.
Existing methods generally follow two technical pipelines, \emph{shadow unaware removal} and \emph{shadow aware removal}, to solve this task.
Despite significant progress, they still suffer challenges in practical applications.
For example, shadow unwary removal methods~\cite{hu2019direction,wang2024softshadow,li2024controlling} lack the ability to perform controllable fine-grained shadow removal, and shadow-unwary removal methods~\cite{guo2023shadowformer,guo2023shadowdiffusion,xiao2024homoformer} rely on precise shadow region masks $\textbf{\emph{m}} \in \mathbb{R}^{h\times w\times 1}$ as a prerequisite. 
These challenges significantly limit the practical application space of shadow removal methods.

In this paper, we propose a new paradigm, \emph{\textbf{prompt-aware controllable shadow removal}}.
It aims to enable removal of shadow from the specified subject based on user prompt $\textbf{\emph{c}}$.
Unlike existing methods, our paradigm can directly learn the mapping $\mathcal{G}$ from $\textbf{\emph{x}}$ to $\widehat{\textbf{\emph{y}}}$ guided by user prompt $\textbf{\emph{c}}$, without requiring any annotations of the shadow region $\textbf{\emph{m}}$.
The proposed prompt-aware controllable shadow removal paradigm can be formulated as follows:
\begin{equation}
\widehat{\textbf{\emph{y}}} = \mathcal{G}(\textbf{\emph{x}}; \textbf{\emph{c}}),
\end{equation}
where $\textbf{\emph{c}}$ can be diverse prompt clues, such as dot, line, subject mask, and so on.
This paradigm not only enables controllable fine-grained shadow removal by user prompts, but also effectively alleviates the need for shadow masks during inference.

\noindent\textbf{{Network Design.}}
In this paper, we design an end-to-end prompt-aware controllable shadow removal network, named PACSRNet.
As illustrated in Fig.~\ref{fig2}, the proposed PACSRNet comprises two key components: a prompt-aware module and a shadow removal module.
First, the prompt-aware module generates a shadow mask and prompt-aware guidance for a specified subject based on user prompts, indicating the content restoration regions for the shadow removal module. 
Then, the shadow removal module aggregates valid contextual information from shadow-free regions using the shadow prior obtained from the prompt-aware module to restore the content of shadow regions.
In our network, the prompt-aware module and shadow removal module are closely correlated and mutually constrained.
The former assists the latter to locate the shadow regions of a specific subject based on user prompt, while the latter regularizes the former by the reconstruction loss, enforcing it focuses on the shadow region of the specified subject.


\subsection{Prompt-Aware Module}
The prompt-aware module is designed to generate a shadow mask and provide prompt-aware guidance of a specified subject based on user prompt, indicating the restoration regions for the shadow removal module.
In the prompt-aware module, it is crucial to locate the shadow boundary of a specific subject based on user prompt.
In fact, there is a significant frequency difference between the shadow and non-shadow regions in an image.
On the one hand, shadow regions tend to exhibit lower frequencies compared to non-shadow ones.
On the other hand, the transition from the shadow region to the non-shadow region is abrupt, creating a distinct gradient change at the shadow boundary.
Consequently, integrating frequency information into spatial features helps the shadow perception module predict the shadow mask.

For this purpose,
we design a spatial-frequency interaction (SFI) block by introducing discrete wavelet transform (DWT) \cite{mallat1989theory}. 
It decomposes the features into frequency components using DWT and interacts with the spatial domain features to enhance the ability to perceive shadow boundaries.
Specifically, for the feature $\textbf{\emph{f}}$ corresponding to the image $\textbf{\emph{x}}$, we first feed it into two different branches to extract the frequency feature $\textbf{\emph{e}}$ and the spatial feature $\textbf{\emph{z}}$, respectively.
\begin{equation}
\textbf{\emph{e}} = \mathcal{D}(\textbf{\emph{f}}),~~
\textbf{\emph{z}} = \mathcal{{C}}_{1}(\textbf{\emph{f}}),
\end{equation}
where $\mathcal{D}(\cdot)$ and $\mathcal{{C}}_{1}(\cdot)$ denote a DWT layer and a $3 \times 3$ convolution layer.
In this way, the feature $\textbf{\emph{e}}$ can capture the frequency variation details in the image, while feature $\textbf{\emph{z}}$ learn the image semantic information in the spatial domain.

After obtained frequency feature $\textbf{\emph{e}}$ and spatial feature $\textbf{\emph{z}}$, we interact the information between them to enhance the representation ability of shadow regions. Such a strategy can better utilize the complementary information between frequency feature $\textbf{\emph{e}}$ and the spatial feature $\textbf{\emph{z}}$ to facilitate shadow perception of the prompt-aware module. 
The whole feature interaction process can be formulated as:
\begin{equation}
\textbf{\emph{g}} = \textbf{\emph{e}}\oplus\mathcal{{C}}_{2}(\textbf{\emph{e}}),~~
\textbf{\emph{h}} = \textbf{\emph{f}}\oplus\mathcal{{C}}_{3}(\textbf{\emph{z}}),
\end{equation}
where $\textbf{\emph{g}}$ and $\textbf{\emph{h}}$ are output of frequency and spatial branches, respectively. $\mathcal{{C}}_{2}(\cdot)$ and $\mathcal{{C}}_{3}(\cdot)$ denote two different $1 \times 1$ convolutional layers. $\oplus$ represents the addition operation.

Finally, the output $\textbf{\emph{g}}$ of the frequency branch is converted into the spatial domain by an inverse DWT layer $\mathcal{I}(\cdot)$ , and aggregated with the features from the spatial branch to obtain the final interaction feature $\textbf{\emph{t}}$. 
\begin{equation}
\textbf{\emph{t}} = \mathcal{A}\big(\mathcal{I}(\textbf{\emph{g}}), \textbf{\emph{h}}\big),
\end{equation}
where $\mathcal{A}(\cdot,\cdot)$ is aggregation convolutional layer. The final predicted shadow mask $\widehat{\textbf{\emph{m}}}$ can be obtained by decoding ${\textbf{\emph{t}}}$ with the decoder. Additionally, we further leverage the feature from prompt-aware module via a linear operation to provide prompt-aware guidance for shadow removal module.

\subsection{Shadow Removal Module}
\label{Shadow-Removal Module}
The shadow removal module aims to restore the content in the predicted shadow regions $\widehat{\textbf{\emph{m}}}$ by aggregating effective contextual information. 
This is consistent with the goal of shadow aware removal settings. 
Recently, benefiting from the advantages of long-range feature capture, transformer-based methods have achieved superior performance.
To enable the local attention mechanism of the transformer to capture global representations, the latest method~\cite{xiao2024homoformer} introduces a random shuffle strategy to enable global interactions.
Although significant improvements have been made, this strategy introduces new challenges.
Specifically, the random shuffle strategy gives each pixel an equal probability of appearing in the same local token.
However, this means that more irrelevant content will be introduced in the local tokens.
In this scenario, aggregating the features using all attention relations based on query-key pairs will introduce redundant or irrelevant content into shadow regions, resulting in blurry or compromised results~\cite{chen2023learning,zhou2024adapt}.
Therefore, we design a dense-sparse local attention (DSLA) block to alleviate the above challenges.

For the local token $\textbf{\emph{l}}_i$ obtained after using the random shuffle strategy, we first map it into query ($\textbf{\emph{q}}_i$), key ($\textbf{\emph{k}}_i$), and value ($\textbf{\emph{v}}_i$) by three different linear layers.
Then, the standard dense-attention score $\textbf{\emph{d}}_i$ between $\textbf{\emph{q}}_i$ and $\textbf{\emph{k}}_i$ can be calculated by a softmax layer:
\begin{equation}
\textbf{\emph{d}}_i={Softmax}\left(\frac{\textbf{\emph{q}}_i\cdot{(\textbf{\emph{k}}_i)^T}}{\sqrt{d}}\right),
\label{SSS1}
\end{equation}
where ${Softmax}(\cdot)$ denotes the softmax layer.
Since the shuffle strategy introduces irrelevant content, Eq.(\ref{SSS1}) will inevitably reduce the attention scores of relevant content, resulting in sub-optimal shadow removal results.

To mitigate the negative impact of irrelevant content, we develop a sparse attention by a screening operation.
The sparse attention score $\textbf{\emph{s}}_i$ can be calculated as follows:
\begin{equation}
\textbf{\emph{s}}_i={Softmax}\left({mask}\left(\frac{\textbf{\emph{q}}_i\cdot{(\textbf{\emph{k}}_i)^T}}{\sqrt{d}}\right)\right),
\label{SSS}
\end{equation}
where $mask$ denote a screening operation. It sets negative similarity values to negative infinity and retains the positive similarity values.

Note that simply relying on $\textbf{\emph{s}}_i$ will result in over sparsity, which in turn causes the encoded features insufficient to restore the shadow region.
Only using $\textbf{\emph{d}}_i$ will inadvertently introduce irrelevant content into shadow regions, leading to distorted and blurry results.
Therefore, we introduce two learnable weights ($\omega_1$ and $\omega_2$) to balance $\textbf{\emph{d}}_i$ and $\textbf{\emph{s}}_i$.
The aggregated feature $\widehat{\textbf{\emph{l}}}_i$ can be calculated as follows:
\begin{equation}
\widehat{\textbf{\emph{l}}}_i=(\omega_1\odot \textbf{\emph{d}}_i+\omega_2\odot \textbf{\emph{s}}_i)\cdot\textbf{\emph{v}}_i,
\label{SSS}
\end{equation}
where $\odot$ denotes the multiply operation. 
Finally, to preserve the original semantic information of the image, all aggregated features will be restored to the original order by the inverse shuffle operation.

Additionally, to obtain more guidance from the prompt-aware module, we connect the multi-scale features extracted by the prompt-aware module to the encoder of the shadow removal module through a linear layer.
In this way, the prompt-aware module can implicitly guide the feature learning of the shadow removal module, enabling controllable shadow removal based on user prompts.

\subsection{Loss Functions}
We train our PACSRNet by minimizing the following loss:
\begin{equation}
\mathcal{L} =  \lambda \mathcal{L}_{re} +\mathcal{L}_{pr},
\end{equation}
where $\mathcal{L}_{re}$ and $\mathcal{L}_{pr}$ denote shadow removal loss and shadow prediction loss, respectively.  
$\lambda$ is a trade-off parameter. 
In real implementation, we empirically set $\lambda=3$. 
\input{tables/tab1}

\section{Experiment}
\subsection{Experiment Setups}
\noindent{\bf{Dataset Generation.}} 
Existing shadow removal datasets \cite{qu2017deshadownet,wang2018STCGAN,le2019shadow} are not suitable for prompt-aware controllable shadow removal task.
On the one hand, these datasets typically include only shadow images, shadow-free images, and shadow masks, lacking diverse user prompts such as a dot, line, and subject mask. 
On the other hand, the image scenes in these datasets typically contain only one shadow region, failing to simulate complex real-world scenes.

In this paper, we customize the prompt-based controllable shadow removal dataset, named PCSRD.
Specifically, we first sample shadow-free images, shadow images, and shadow masks with multiple subject scenes from DESOBAv2~\cite{liu2024shadow} dataset, and collect the corresponding subject mask as one of the user prompts.
Subsequently, to obtain diverse user prompts, we employ the subject mask to automatically generate the corresponding dot and line prompts by a dynamic programming strategy.
In this way, customized PCSRD dataset can effectively simulate complex real-world scenes with multiple subjects and diverse prompts.

The final dataset consists of 11,900 complex scenes samples, which are randomly divided into 10,000 for training, 1,000 for validation, and 900 for testing.
To ensure the validity of the results, we carefully prevented any data leakage between the training and testing sets.
To the best of our knowledge, this is the first dataset for prompt-aware controllable shadow removal task. 
The dataset will be published to facilitate subsequent research.

\noindent\textbf{Testing Dataset.}
In addition to our customized PCSRD dataset, we also introduce the ISTD+~\cite{le2019shadow} dataset to validate the effectiveness of our shadow removal module.
ISTD+~\cite{le2019shadow} dataset includes 1,330 training and 540 testing triplets.
Each triplet consists of a shadow image, shadow-free image, and shadow region mask.
The ISTD+ dataset is obtained by reducing the illumination inconsistency between the shadow and shadow-free image of ISTD~\cite{wang2018STCGAN} dataset.



\noindent{\bf{Baselines.}}
To the best of our knowledge, there is no existing work focusing on prompt-based controllable shadow removal task. 
Therefore, we use five state-of-the-art shadow removal methods as our baselines to evaluate the performance of our PACSRNet, including BMNet~\cite{zhu2022bijective}, ShadowFormer~\cite{guo2023shadowformer}, ShadowDiffusion~\cite{guo2023shadowdiffusion}, Inpaint4Shadow~\cite{li2023leveraging}, and HomoFormer~\cite{xiao2024homoformer}.
To ensure the comparability of experimental results, these baseline methods used for comparison are fine-tuned on our PCSRD dataset through their released models and codes.


\noindent{\bf{Evaluation Metrics.}} 
Following previous works~\cite{wang2018STCGAN,qu2017deshadownet,le2019shadow}, we employ the peak signal-to-noise ratio (PSNR)~\cite{wu2023flow}, structural similarity (SSIM)~\cite{10447901}, and root mean square error (RMSE)~\cite{9446636} as quantitative evaluation metrics. 
Specifically, PSNR and SSIM are two popular metrics to evaluate image fidelity of shadow removal results.
The higher value ($\uparrow$) of PSNR and SSIM metrics indicates better performance.
RMSE measures the mean error between the pixels of the results and the ground truth.
The lower value ($\downarrow$) of RMSE indicates better shadow removal.
Furthermore, to perform detailed analysis of the removal effects, we calculate all the above metrics for shadow regions, non-shadow regions, and all regions, comparing the ground truth with the generated removal results for each region separately.

\begin{figure*}[htbp]
\centering
\includegraphics[width=1.0\linewidth]{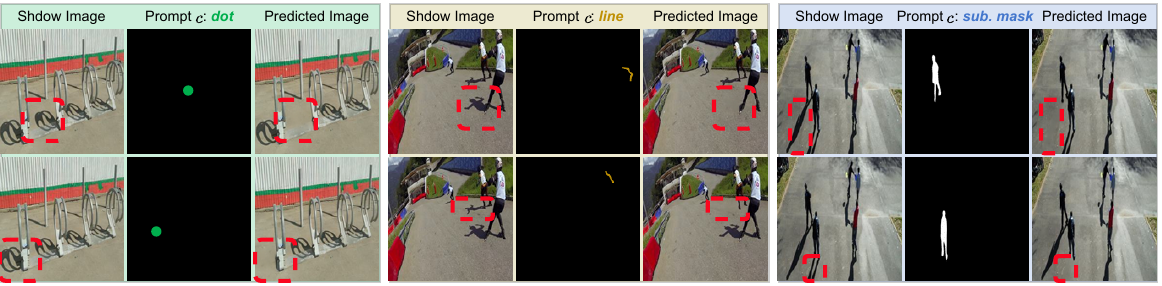}
\vspace{-0.4cm}
\caption{Examples of shadow removal results based on dot, line and subject mask prompts. In the same image, we use the prompt to specify the different subjects and perform corresponding controllable shadow removal.}
\label{fig:vis_pred}
\vspace{-0.1cm}
\end{figure*}

\subsection{Experimental Results and Analysis}
\noindent\textbf{Quantitative Results.} 
The quantitative results of shadow removal are reported in Tab.~\ref{tab:results_pcsrd}. 
As shown in the table, our method achieves performance comparable to ShadowDiffusion~\cite{guo2023shadowdiffusion} in terms of PSNR across the entire image under three different user prompts.
Specifically, the PSNR of our method is 40.95 dB, 41.06 dB, and 41.59 dB under point, line, and subject mask prompts, respectively.
These results outperform the PSNR of ShadowDiffusion (40.82 dB)~\cite{guo2023shadowdiffusion}, which requires precise shadow region masks as priors.
This demonstrates the effectiveness of the proposed prompt-aware controllable shadow removal framework.

\noindent \textbf{Qualitative Evaluation.}
To visually validate the effectiveness of the proposed PACSRNet, we present the shadow removal results of PACSRNet under three different prompts in Fig.~\ref{fig:vis_pred}.
Specifically,  the left of Fig.~\ref{fig:vis_pred} shows an example of shadow removal based on a dot prompt, the middle displays the shadow removal case based on a line prompt, and the right demonstrates the shadow removal case based on a subject mask prompt.
From Fig.~\ref{fig:vis_pred}, we can clearly observe that the proposed PACSRNet is capable of effectively removing the shadows of the user-specified subject under all three types of prompts. 
These results highlight the flexibility and robustness of PACSRNet in effectively handling a wide range of user-specified prompts, demonstrating its ability to perform high-quality shadow removal across different scenarios. 
In addition, we can see in Fig.~\ref{fig:vis_pred} that PACSRNet can remove shadows of different subjects using the dot, line, and subject mask prompts.
For instance, in the example of the point prompt, PACSRNet effectively removes the shadows from different subjects by processing prompt placed at various locations across the image. 
This flexibility further emphasizes the robustness of PACSRNet in controllable shadow removal.

\input{tables/tab3}

\subsection{Ablation Study}
\noindent{\bf{Effectiveness of Spatial-Frequency Interaction (SFI)}}. 
We conduct ablation study to verify the effectiveness of SFI block.
Specifically, we compare the full model with SFI block and the model without SFI block in Tab.~\ref{tab3}.
From the results, we can observe that SFI block significantly improve the performance of the model. 
Its PSNR value increases by 0.2326 dB.
These results illustrate that frequency features are beneficial for the prediction and restoration of shadow regions.



\begin{figure}[t!]
\centering
\includegraphics[width=1.0\linewidth]{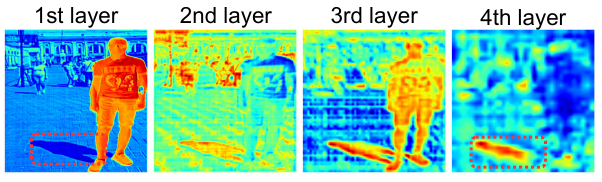}
\caption{Visualization of prompt-aware guidance feature maps. It implicitly guides the shadow removal module to focus on shadow regions marked in red dashed box.}
\label{fig:vis_heatmap}
\end{figure}

\input{tables/tab5}

\noindent{\bf{Effectiveness of Dense-Sparse Local Attention (DSLA).}} 
In Tab. \ref{tab3}, we conduct an ablation study to validate the effectiveness of DSLA. 
From the table, we can observe that the standard dense attention model without sparse branch achieves the worst removal performance due to the introduction of irrelevant content. 
Its PSNR is improved by 0.11 dB compared to the dense-sparse local attention model with sparse attention branch.
These results demonstrate that using sparse attention branch to mitigate the negative impact of irrelevant content on the attention computation is effective.


\noindent{\bf{Effectiveness of Prompt-Aware Guidance Strategy.}}
As mentioned in the section~\ref{Shadow-Removal Module}, we connect the multi-scale features from the prompt-aware module to the encoder of the shadow removal module, implicitly guiding its feature learning.
To verify the effectiveness of the prompt-aware guidance strategy, we compare the full model with those without the prompt-aware guidance strategy.
As shown in Tab.~\ref{tab3}, the full model using the prompt-aware guidance strategy achieves better shadow removal performance.
Furthermore, 
we visualize the multi-scale feature maps in the prompt-aware module used to guide the feature learning of the shadow removal module in Fig.~\ref{fig:vis_heatmap}.
As shown in Fig.~\ref{fig:vis_heatmap}, we can see that the features at different scales guide different aspects of the shadow removal module.
For example, low-level features focus on the subject, while high-level features emphasize the shadow regions related to the subject.
These results verify the necessity and effectiveness of the prompt-aware guidance strategy.

\begin{figure}[!t]
\centering
\includegraphics[width=1.01\linewidth]{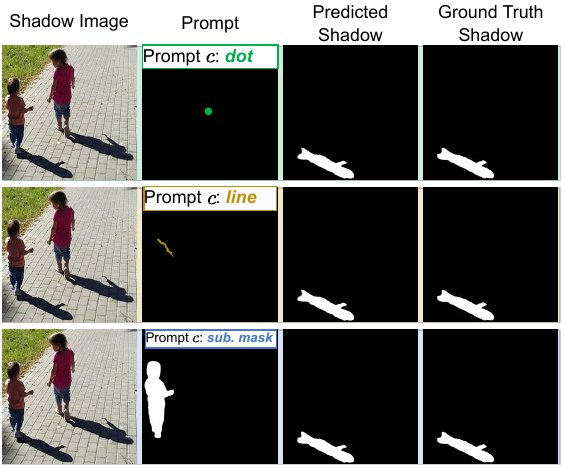}
\caption{Visual comparison of the predicted shadow region masks for one subject under three different prompts. Our PACSRNet is robust to different types of user prompts.}
\label{figp}
\end{figure}

\noindent{\bf{Robustness to User Prompts.}}
In our design, the proposed PACSRNet supports various user prompts, such as dots, lines, and subject masks. 
Here, we verify its robustness to user prompts. 
As shown in Tab.~\ref{tab5}, we compare the shadow region masks predicted by PACSRNet under three different prompts.
From the table, we can observe that the three different prompts yield similar intersection over union (IoU) and binary cross entropy (BCE) values.
In addition, we also present the shadow region prediction results of PACSRNet under three different user prompts.
As shown in Fig.~\ref{figp}, PACSRNet obtains shadow region masks consistent with ground truth under three different user prompts.
These results further demonstrate that PACSRNet is robust to user prompts.

\begin{figure}[t!]
\centering
\includegraphics[width=1\linewidth]{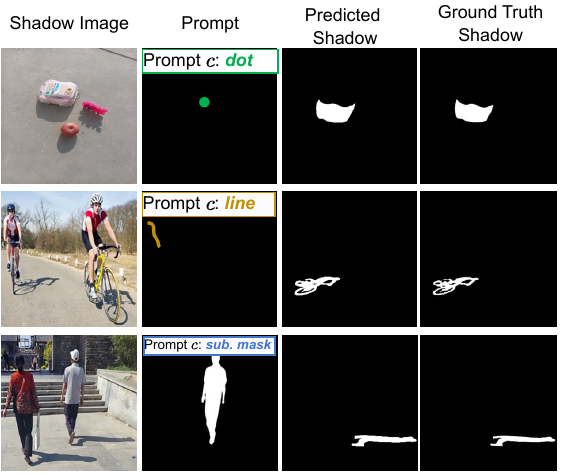}
\caption{Visual examples of our PACSRNet predicting shadow regions under three different prompts.}
\label{fig:example2}
\end{figure}

\noindent{\bf{Effectiveness of Prompt-Aware Module.}} 
In Fig.~\ref{fig:example2}, we visualize the shadow region prediction results of the prompt-aware module under three different prompts.
From the figure, we can see that the prompt-aware module can predict the shadow mask of the user-specified subject among multiple subjects under three different prompts.
For example, in the first case, the prompt-aware module predicts the shadow regions of the schoolbag based on the given dot prompt.
These results demonstrate the effectiveness and superiority of the designed prompt-aware module in shadow prediction.

\begin{figure}[t!]
\centering
\includegraphics[width=1\linewidth]{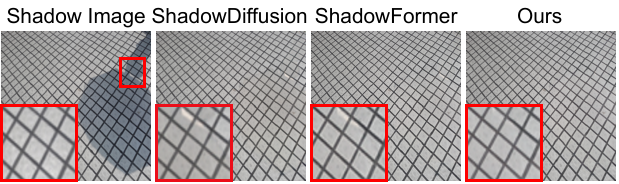}
\caption{Example of shadow removal results on the ISTD+~\protect\cite{le2019shadow} dataset. The input shadow image, the estimated results of ShadowDiffusion~\protect\cite{guo2023shadowdiffusion}, ShadowFormer~\protect\cite{guo2023shadowformer}, and ours, respectively. The slice of the shadow image corresponds to the ground truth shadow-free image.}
\label{fig:example3}
\end{figure}

\noindent{\bf{Effectiveness of Shadow Removal Module.}} 
To further verify the effectiveness of the shadow removal module, we compare it with five baselines on two datasets (PCSRD and ISTD +~\cite{le2019shadow}).
As shown in Tab.~\ref{tab:results_pcsrd}, our shadow removal module achieves superior performance on both PCSRD and ISTD+~\cite{le2019shadow} datasets. 
In particular, for the shadow regions, our shadow removal module outperforms the five baselines by a large margin on both the PCSRD and ISTD +~\cite{le2019shadow} datasets. 
The specific earnings of it are 0.116 dB and 0.63 dB in shadow regions on both PCSRD and ISTD+~\cite{le2019shadow} datasets, respectively.
Furthermore, Fig.~\ref{fig:example3} presents an example comparing our shadow removal module with two competitive baselines: ShadowFormer~\cite{guo2023shadowformer} and ShadowDiffusion~\cite{guo2023shadowdiffusion}.
As can be observed, the shadow regions restored by our shadow removal module are more plausible.
These results demonstrate the superiority of our shadow removal module in shadow removal.



\section{Conclusion}
This paper develops a prompt-aware controllable shadow removal network, which consists of two key components: a prompt-aware module and a shadow removal module.
The former aims to generate the shadow masks for a specified subject based on user prompts (\emph{e.g.}, dot, line, and subject mask), while the latter aggregates relevant content by a dense-sparse local attention block to restore the shadow regions predicted by the former.
Such a design not only achieves controllable shadow removal, but also avoids the non-trivial and tedious annotations of shadow region.
Furthermore, we introduce the first dataset suitable for the prompt-aware controllable shadow removal task, which can effectively facilitate subsequent research.
Extensive experimental results demonstrate the superiority and flexibility of our method.

\bibliographystyle{named}
\bibliography{ijcai25}

\end{document}

%% file: tables/tab1.tex
\begin{table*}[!t]
\vspace{-0.2cm}
\centering
\tabcolsep 10pt
\renewcommand\arraystretch{1.1}
\centering
\resizebox{1.0\linewidth}{!}{
\begin{tabular}{|l||c|c|c|c|c|c|c|c|c|}
\hline\thickhline
\rowcolor[HTML]{f8f9fa}   & \multicolumn{3}{c|}{\textbf{Shadow Regions}}      & \multicolumn{3}{c|}{\textbf{Non-Shadow Regions}}  & \multicolumn{3}{c|}{\textbf{All Regions}}\\
 \cline{2-10}  
\rowcolor[HTML]{f8f9fa} \multirow{-2}{*}{ \textbf{Methods}}    &  PSNR$\uparrow$ &  SSIM$\uparrow$ &   RMSE$\downarrow$  &  PSNR$\uparrow$ &  SSIM$\uparrow$ &   RMSE$\downarrow$ &  PSNR$\uparrow$ &  SSIM$\uparrow$ &   RMSE$\downarrow$  \\
\hline
\multicolumn{10}{|c|}{PCSRD dataset} \\ \hline
BMNet~\cite{zhu2022bijective}   & 44.459 & 0.9955 & 5.9428 & 48.832 & 0.9961 & 0.6401 & 41.627 & 0.9881 & 0.7743 \\
Inpaint4Shadow~\cite{li2023leveraging} & \underline{45.443} & \underline{0.9958} & 8.6692 & 45.643 & 0.9955 & 0.8817 & 41.969 & 0.9884 & 0.9723 \\
ShadowFormer~\cite{guo2023shadowformer}   & 45.318 & 0.9957 & \underline{5.1687} & \underline{49.774} & 0.9962 & \underline{0.5018} & \underline{42.302} & \underline{0.9889} & \underline{0.6183} \\
ShadowDiffusion~\cite{guo2023shadowdiffusion} & 44.593 & 0.9896 & 12.507 & 46.094 & 0.9889 & 1.0726 & 40.823 & 0.9869 & 1.2135 \\
HomoFormer~\cite{xiao2024homoformer} & 45.256 & 0.9957 & 5.3159 & 49.593 & \underline{0.9962} & 0.5042 & 42.219 & 0.9888 & 0.6251 \\
\rowcolor[HTML]{DAE8FC} \textbf{PACSRNet$^\dagger$ (Ours)}   & \textbf{45.559} & \textbf{0.9959} & \textbf{4.9987} & \textbf{49.784} & \textbf{0.9964} & \textbf{0.4927} & \textbf{42.494} & \textbf{0.9892} & \textbf{0.6038} \\ \hline
\rowcolor[HTML]{f8f9fa} \textbf{PACSRNet (Ours) w/ Dot }   & {43.382} & {0.9952} & {6.1311} & {48.622} & {0.9961} & {0.4864} & {40.956} & {0.9878} & {0.6341} \\
\rowcolor[HTML]{f8f9fa} \textbf{PACSRNet (Ours) w/ Line }   & {43.479} & {0.9953} & {6.0847} & {48.867} & {0.9961} & {0.4861} & {41.056} & {0.9879} & {0.6333} \\
\rowcolor[HTML]{f8f9fa} \textbf{PACSRNet (Ours) w/ Subject Mask}  & {44.354} & {0.9957} & {5.5541} & {49.061} & {0.9961} & {0.5022} & {41.592} & {0.9884} & {0.6263}\\
\hline
\multicolumn{10}{|c|}{ISTD+ dataset} \\ \hline
BMNet~\cite{zhu2022bijective} & 37.87 & 0.991 & 5.62 & 37.51 & 0.985 & 2.45 & 33.98 & 0.972 & 2.97 \\
Inpaint4Shadow~\cite{li2023leveraging} & 38.10 & 0.990 & 6.09 & 37.66 & 0.981 & 2.82 & 34.16 & 0.967 & 3.35 \\ 
ShadowFormer~\cite{guo2023shadowformer} & 39.48 & \underline{0.992} & 5.23 & 38.82 & 0.983 & 2.30 & 35.46 & 0.971 & 2.78 \\
ShadowDiffusion~\cite{guo2023shadowdiffusion} & \underline{39.69} & \underline{0.992} & 4.97 & 38.89 & \textbf{0.987} & 2.28 & \underline{35.67} & \textbf{0.975} & 2.72 \\
HomoFormer~\cite{xiao2024homoformer} & 39.49 & \textbf{0.993} & \textbf{4.73} & 38.75 & 0.984 & \textbf{2.23} & 35.35 & \textbf{0.975} & \underline{2.64} \\ \hline
\rowcolor[HTML]{DAE8FC} \textbf{PACSRNet$^\dagger$ (Ours)} & \textbf{40.32} & \textbf{0.993} & \underline{4.89} & \textbf{39.18} & \underline{0.985} & \underline{2.27} & \textbf{36.02} & \underline{0.972} & \textbf{2.63} \\ 
\hline

\end{tabular}}
\vspace{-0.2cm}
\caption{Comparison with the state-of-the-art methods on the PCSRD and ISTD+~\protect\cite{le2019shadow} datasets. 
PACSRNet$^\dagger$ denotes the model only uses the shadow removal module. 
The best and second best results are \textbf{boldfaced} and \underline{underlined}.
}
\label{tab:results_pcsrd}
\vspace{-0.2cm}
\end{table*}

%% file: tables/tab3.tex
\begin{table}[t!] 
\tabcolsep 8pt
\renewcommand\arraystretch{1.1}
\resizebox{1.0\linewidth}{!}{
\begin{tabular}{|l||c|c|c|}
\hline\thickhline
\rowcolor[HTML]{f8f9fa} Method & PSNR $\uparrow$& SSIM $\uparrow$& RMSE $\downarrow$ \\ \hline
PACSRNet w/o SFI & 42.265 & 0.9888 & 0.6330  \\
PACSRNet w/o LDSA & 42.384 & 0.9890 & 0.6075 \\
PACSRNet w/o Guidance & 41.215 & 0.9881 & 0.6262 \\ \hline
\rowcolor[HTML]{DAE8FC} PACSRNet (Full) & \textbf{42.494} & \textbf{0.9892} & \textbf{0.6038}  \\ \hline
\end{tabular}}
\vspace{-0.2cm}
\caption{Ablation results for the key blocks of PSCSRNet.}
\label{tab3}
\vspace{-0.2cm}
\end{table}

%% file: tables/tab5.tex
\begin{table}[t!] 
\centering
\tabcolsep 19pt
\renewcommand\arraystretch{1.1}
\resizebox{1.0\linewidth}{!}{
\begin{tabular}{|l||c|c|}
\hline\thickhline
\rowcolor[HTML]{f8f9fa} Method & IoU $\uparrow$ & BCE $\downarrow$ \\ \hline
PACSRNet - dot  & 0.816 & 0.018    \\ \hline
PACSRNet - line   & 0.825 & 0.016    \\ \hline
PACSRNet - subject mask & 0.872 & 0.013    \\ \hline
\end{tabular}}
\vspace{-0.1cm}
\caption{Evaluation of shadow masks predicted using different types of prompts in PSCSRNet.}
\label{tab5}
\vspace{-1.5mm}
\end{table}

%% file: main.bbl
\begin{thebibliography}{}

\bibitem[\protect\citeauthoryear{Alaluf \bgroup \em et al.\egroup }{2022}]{alaluf2022hyperstyle}
Yuval Alaluf, Omer Tov, Ron Mokady, Rinon Gal, and Amit Bermano.
\newblock Hyperstyle: Stylegan inversion with hypernetworks for real image editing.
\newblock In {\em Proc. CVPR}, pages 18511--18521, 2022.

\bibitem[\protect\citeauthoryear{Chen \bgroup \em et al.\egroup }{2020}]{chen2020multi}
Zhihao Chen, Lei Zhu, Liang Wan, Song Wang, Wei Feng, and Pheng-Ann Heng.
\newblock A multi-task mean teacher for semi-supervised shadow detection.
\newblock In {\em Proc. CVPR}, pages 5611--5620, 2020.

\bibitem[\protect\citeauthoryear{Chen \bgroup \em et al.\egroup }{2023}]{chen2023learning}
Xiang Chen, Hao Li, Mingqiang Li, and Jinshan Pan.
\newblock Learning a sparse transformer network for effective image deraining.
\newblock In {\em Proc. CVPR}, pages 5896--5905, 2023.

\bibitem[\protect\citeauthoryear{Du \bgroup \em et al.\egroup }{2022}]{du2022elements}
Hang Du, Hailin Shi, Dan Zeng, Xiao-Ping Zhang, and Tao Mei.
\newblock The elements of end-to-end deep face recognition: A survey of recent advances.
\newblock {\em ACM CSUR}, 54(10s):1--42, 2022.

\bibitem[\protect\citeauthoryear{Finlayson \bgroup \em et al.\egroup }{2005}]{finlayson2005removal}
Graham~D Finlayson, Steven~D Hordley, Cheng Lu, and Mark~S Drew.
\newblock On the removal of shadows from images.
\newblock {\em IEEE TPAMI}, 28(1):59--68, 2005.

\bibitem[\protect\citeauthoryear{Finlayson \bgroup \em et al.\egroup }{2009}]{finlayson2009entropy}
Graham~D Finlayson, Mark~S Drew, and Cheng Lu.
\newblock Entropy minimization for shadow removal.
\newblock {\em IJCV}, 85(1):35--57, 2009.

\bibitem[\protect\citeauthoryear{Guo \bgroup \em et al.\egroup }{2023a}]{guo2023shadowformer}
Lanqing Guo, Siyu Huang, Ding Liu, Hao Cheng, and Bihan Wen.
\newblock Shadowformer: Global context helps image shadow removal.
\newblock {\em arXiv preprint arXiv:2302.01650}, 2023.

\bibitem[\protect\citeauthoryear{Guo \bgroup \em et al.\egroup }{2023b}]{guo2023shadowdiffusion}
Lanqing Guo, Chong Wang, Wenhan Yang, Siyu Huang, Yufei Wang, Hanspeter Pfister, and Bihan Wen.
\newblock Shadowdiffusion: When degradation prior meets diffusion model for shadow removal.
\newblock In {\em Proc. CVPR}, pages 14049--14058, 2023.

\bibitem[\protect\citeauthoryear{Hu \bgroup \em et al.\egroup }{2019}]{hu2019direction}
Xiaowei Hu, Chi-Wing Fu, Lei Zhu, Jing Qin, and Pheng-Ann Heng.
\newblock Direction-aware spatial context features for shadow detection and removal.
\newblock {\em IEEE TPAMI}, 42(11):2795--2808, 2019.

\bibitem[\protect\citeauthoryear{Huang \bgroup \em et al.\egroup }{2011}]{huang2011characterizes}
Xiang Huang, Gang Hua, Jack Tumblin, and Lance Williams.
\newblock What characterizes a shadow boundary under the sun and sky?
\newblock In {\em Proc. ICCV}, pages 898--905. IEEE, 2011.

\bibitem[\protect\citeauthoryear{Khan \bgroup \em et al.\egroup }{2014}]{khan2014automatic}
Salman~Hameed Khan, Mohammed Bennamoun, Ferdous Sohel, and Roberto Togneri.
\newblock Automatic feature learning for robust shadow detection.
\newblock In {\em Proc. CVPR}, pages 1939--1946. IEEE, 2014.

\bibitem[\protect\citeauthoryear{Kim \bgroup \em et al.\egroup }{2022}]{kim2022adaface}
Minchul Kim, Anil~K Jain, and Xiaoming Liu.
\newblock Adaface: Quality adaptive margin for face recognition.
\newblock In {\em Proc. CVPR}, pages 18750--18759, 2022.

\bibitem[\protect\citeauthoryear{Le and Samaras}{2019}]{le2019shadow}
Hieu Le and Dimitris Samaras.
\newblock Shadow removal via shadow image decomposition.
\newblock In {\em Proc. ICCV}, pages 8578--8587, 2019.

\bibitem[\protect\citeauthoryear{Le \bgroup \em et al.\egroup }{2018}]{le2018a+}
Hieu Le, Tomas F~Yago Vicente, Vu~Nguyen, Minh Hoai, and Dimitris Samaras.
\newblock A+ d net: Training a shadow detector with adversarial shadow attenuation.
\newblock In {\em Proc. ECCV}, pages 662--678, 2018.

\bibitem[\protect\citeauthoryear{Li \bgroup \em et al.\egroup }{2023}]{li2023leveraging}
Xiaoguang Li, Qing Guo, Rabab Abdelfattah, Di~Lin, Wei Feng, Ivor Tsang, and Song Wang.
\newblock Leveraging inpainting for single-image shadow removal.
\newblock In {\em Proc. ICCV}, pages 13055--13064, 2023.

\bibitem[\protect\citeauthoryear{Li \bgroup \em et al.\egroup }{2024}]{li2024controlling}
Xinjie Li, Yang Zhao, Dong Wang, Yuan Chen, Li~Cao, and Xiaoping Liu.
\newblock Controlling the latent diffusion model for generative image shadow removal via residual generation.
\newblock {\em arXiv preprint arXiv:2412.02322}, 2024.

\bibitem[\protect\citeauthoryear{Liao \bgroup \em et al.\egroup }{2021}]{liao2021shadow}
Jingwei Liao, Yanli Liu, Guanyu Xing, Housheng Wei, Jueyu Chen, and Songhua Xu.
\newblock Shadow detection via predicting the confidence maps of shadow detection methods.
\newblock In {\em Proc. ACMMM}, pages 704--712, 2021.

\bibitem[\protect\citeauthoryear{Liu \bgroup \em et al.\egroup }{2024a}]{liu2024shadow}
Qingyang Liu, Junqi You, Jianting Wang, Xinhao Tao, Bo~Zhang, and Li~Niu.
\newblock Shadow generation for composite image using diffusion model.
\newblock In {\em Proc. CVPR}, pages 8121--8130, 2024.

\bibitem[\protect\citeauthoryear{Liu \bgroup \em et al.\egroup }{2024b}]{liu2024recasting}
Yuhao Liu, Zhanghan Ke, Ke~Xu, Fang Liu, Zhenwei Wang, and Rynson~WH Lau.
\newblock Recasting regional lighting for shadow removal.
\newblock In {\em Proc. AAAI}, volume~38, pages 3810--3818, 2024.

\bibitem[\protect\citeauthoryear{Luo \bgroup \em et al.\egroup }{2024}]{luo2024soc}
Zhuoyan Luo, Yicheng Xiao, Yong Liu, Shuyan Li, Yitong Wang, Yansong Tang, Xiu Li, and Yujiu Yang.
\newblock Soc: Semantic-assisted object cluster for referring video object segmentation.
\newblock {\em Proc. NeurIPS}, 36, 2024.

\bibitem[\protect\citeauthoryear{Mallat}{1989}]{mallat1989theory}
Stephane~G Mallat.
\newblock A theory for multiresolution signal decomposition: the wavelet representation.
\newblock {\em IEEE TPAMI}, 11(7):674--693, 1989.

\bibitem[\protect\citeauthoryear{Mei \bgroup \em et al.\egroup }{2024}]{mei2024latent}
Kangfu Mei, Luis Figueroa, Zhe Lin, Zhihong Ding, Scott Cohen, and Vishal~M Patel.
\newblock Latent feature-guided diffusion models for shadow removal.
\newblock In {\em Proc. WACV}, pages 4313--4322, 2024.

\bibitem[\protect\citeauthoryear{Meinhardt \bgroup \em et al.\egroup }{2022}]{meinhardt2022trackformer}
Tim Meinhardt, Alexander Kirillov, Laura Leal-Taixe, and Christoph Feichtenhofer.
\newblock Trackformer: Multi-object tracking with transformers.
\newblock In {\em Proc. CVPR}, pages 8844--8854, 2022.

\bibitem[\protect\citeauthoryear{Qu \bgroup \em et al.\egroup }{2017}]{qu2017deshadownet}
Liangqiong Qu, Jiandong Tian, Shengfeng He, Yandong Tang, and Rynson~WH Lau.
\newblock Deshadownet: A multi-context embedding deep network for shadow removal.
\newblock In {\em Proc. CVPR}, pages 4067--4075, 2017.

\bibitem[\protect\citeauthoryear{Shen \bgroup \em et al.\egroup }{2015}]{shen2015shadow}
Li~Shen, Teck Wee~Chua, and Karianto Leman.
\newblock Shadow optimization from structured deep edge detection.
\newblock In {\em Proc. CVPR}, pages 2067--2074, 2015.

\bibitem[\protect\citeauthoryear{Shor and Lischinski}{2008}]{shor2008shadow}
Yael Shor and Dani Lischinski.
\newblock The shadow meets the mask: Pyramid-based shadow removal.
\newblock In {\em Computer Graphics Forum}, volume~27, pages 577--586. Wiley Online Library, 2008.

\bibitem[\protect\citeauthoryear{Vasluianu \bgroup \em et al.\egroup }{2024}]{vasluianu2024ntire}
Florin-Alexandru Vasluianu, Tim Seizinger, Zhuyun Zhou, Zongwei Wu, Cailian Chen, Radu Timofte, Wei Dong, Han Zhou, Yuqiong Tian, Jun Chen, et~al.
\newblock Ntire 2024 image shadow removal challenge report.
\newblock In {\em Proc. CVPR}, pages 6547--6570, 2024.

\bibitem[\protect\citeauthoryear{Vicente \bgroup \em et al.\egroup }{2015}]{vicente2015leave}
Tom{\'a}s F~Yago Vicente, Minh Hoai, and Dimitris Samaras.
\newblock Leave-one-out kernel optimization for shadow detection.
\newblock In {\em Proc. ICCV}, pages 3388--3396, 2015.

\bibitem[\protect\citeauthoryear{Vicente \bgroup \em et al.\egroup }{2017}]{vicente2017leave}
Tomas F~Yago Vicente, Minh Hoai, and Dimitris Samaras.
\newblock Leave-one-out kernel optimization for shadow detection and removal.
\newblock {\em IEEE TPAMI}, 40(3):682--695, 2017.

\bibitem[\protect\citeauthoryear{Wang \bgroup \em et al.\egroup }{2018}]{wang2018STCGAN}
Jifeng Wang, Xiang Li, and Jian Yang.
\newblock Stacked conditional generative adversarial networks for jointly learning shadow detection and shadow removal.
\newblock In {\em Proc. CVPR}, 2018.

\bibitem[\protect\citeauthoryear{Wang \bgroup \em et al.\egroup }{2020}]{wang2020instance}
Tianyu Wang, Xiaowei Hu, Qiong Wang, Pheng-Ann Heng, and Chi-Wing Fu.
\newblock Instance shadow detection.
\newblock In {\em Proc. CVPR}, pages 1880--1889, 2020.

\bibitem[\protect\citeauthoryear{Wang \bgroup \em et al.\egroup }{2021}]{wang2021single}
Tianyu Wang, Xiaowei Hu, Chi-Wing Fu, and Pheng-Ann Heng.
\newblock Single-stage instance shadow detection with bidirectional relation learning.
\newblock In {\em Proc. CVPR}, pages 1--11, 2021.

\bibitem[\protect\citeauthoryear{Wang \bgroup \em et al.\egroup }{2022}]{wang2022instance}
Tianyu Wang, Xiaowei Hu, Pheng-Ann Heng, and Chi-Wing Fu.
\newblock Instance shadow detection with a single-stage detector.
\newblock {\em IEEE TPAMI}, 45(3):3259--3273, 2022.

\bibitem[\protect\citeauthoryear{Wang \bgroup \em et al.\egroup }{2024a}]{10447901}
Jianan Wang, Zhiliang Wu, Hanyu Xuan, and Yan Yan.
\newblock Text-video completion networks with motion compensation and attention aggregation.
\newblock In {\em Proc. ICASSP}, pages 2990--2994, 2024.

\bibitem[\protect\citeauthoryear{Wang \bgroup \em et al.\egroup }{2024b}]{wang2024softshadow}
Xinrui Wang, Lanqing Guo, Xiyu Wang, Siyu Huang, and Bihan Wen.
\newblock Softshadow: Leveraging penumbra-aware soft masks for shadow removal.
\newblock {\em arXiv preprint arXiv:2409.07041}, 2024.

\bibitem[\protect\citeauthoryear{Wu \bgroup \em et al.\egroup }{2021}]{9446636}
Zhiliang Wu, Kang Zhang, Hanyu Xuan, Jian Yang, and Yan Yan.
\newblock Dapc-net: Deformable alignment and pyramid context completion networks for video inpainting.
\newblock {\em SPL}, 28:1145--1149, 2021.

\bibitem[\protect\citeauthoryear{Wu \bgroup \em et al.\egroup }{2023}]{wu2023flow}
Zhiliang Wu, Kang Zhang, Changchang Sun, Hanyu Xuan, and Yan Yan.
\newblock Flow-guided deformable alignment network with self-supervision for video inpainting.
\newblock In {\em Proc. ICASSP}, pages 1--5, 2023.

\bibitem[\protect\citeauthoryear{Wu \bgroup \em et al.\egroup }{2024a}]{wu2024waveformer}
Zhiliang Wu, Changchang Sun, Hanyu Xuan, Gaowen Liu, and Yan Yan.
\newblock Waveformer: Wavelet transformer for noise-robust video inpainting.
\newblock In {\em Proc. AAAI}, volume~38, pages 6180--6188, 2024.

\bibitem[\protect\citeauthoryear{Wu \bgroup \em et al.\egroup }{2024b}]{wu2024single}
Zongwei Wu, Jilai Zheng, Xiangxuan Ren, Florin-Alexandru Vasluianu, Chao Ma, Danda~Pani Paudel, Luc Van~Gool, and Radu Timofte.
\newblock Single-model and any-modality for video object tracking.
\newblock In {\em Proc. CVPR}, pages 19156--19166, 2024.

\bibitem[\protect\citeauthoryear{Xiao \bgroup \em et al.\egroup }{2024}]{xiao2024homoformer}
Jie Xiao, Xueyang Fu, Yurui Zhu, Dong Li, Jie Huang, Kai Zhu, and Zheng-Jun Zha.
\newblock Homoformer: Homogenized transformer for image shadow removal.
\newblock In {\em Proc. CVPR}, pages 25617--25626, 2024.

\bibitem[\protect\citeauthoryear{Yang and Yang}{2022}]{yang2022decoupling}
Zongxin Yang and Yi~Yang.
\newblock Decoupling features in hierarchical propagation for video object segmentation.
\newblock {\em Proc. NeurIPS}, 35:36324--36336, 2022.

\bibitem[\protect\citeauthoryear{Yang \bgroup \em et al.\egroup }{2012}]{yang2012shadow}
Qingxiong Yang, Kar-Han Tan, and Narendra Ahuja.
\newblock Shadow removal using bilateral filtering.
\newblock {\em IEEE TIP}, 21(10):4361--4368, 2012.

\bibitem[\protect\citeauthoryear{Yang \bgroup \em et al.\egroup }{2023}]{yang2023silt}
Han Yang, Tianyu Wang, Xiaowei Hu, and Chi-Wing Fu.
\newblock Silt: Shadow-aware iterative label tuning for learning to detect shadows from noisy labels.
\newblock In {\em Proc. ICCV}, pages 12687--12698, 2023.

\bibitem[\protect\citeauthoryear{Zhang \bgroup \em et al.\egroup }{2023}]{10222097}
Yanni Zhang, Zhiliang Wu, and Yan Yan.
\newblock Pfta-net: Progressive feature alignment and temporal attention fusion networks for video inpainting.
\newblock In {\em Proc. ICIP}, pages 191--195, 2023.

\bibitem[\protect\citeauthoryear{Zhang \bgroup \em et al.\egroup }{2024}]{zhang2024advances}
Zixuan Zhang, Xinge Guo, and Chengkuo Lee.
\newblock Advances in olfactory augmented virtual reality towards future metaverse applications.
\newblock {\em NC}, 15(1):6465, 2024.

\bibitem[\protect\citeauthoryear{Zhou \bgroup \em et al.\egroup }{2024}]{zhou2024adapt}
Shihao Zhou, Duosheng Chen, Jinshan Pan, Jinglei Shi, and Jufeng Yang.
\newblock Adapt or perish: Adaptive sparse transformer with attentive feature refinement for image restoration.
\newblock In {\em Proc. CVPR}, pages 2952--2963, 2024.

\bibitem[\protect\citeauthoryear{Zhu \bgroup \em et al.\egroup }{2022}]{zhu2022bijective}
Yurui Zhu, Jie Huang, Xueyang Fu, Feng Zhao, Qibin Sun, and Zheng-Jun Zha.
\newblock Bijective mapping network for shadow removal.
\newblock In {\em Proc. CVPR}, pages 5627--5636, 2022.

\end{thebibliography}
